\documentclass[11pt,a4paper]{article}
\usepackage[hyperref]{emnlp-ijcnlp-2019}
\usepackage{times}
\usepackage{latexsym}

\usepackage{url}
\usepackage[draft,textsize=footnotesize]{todonotes}
\usepackage{graphicx}
\usepackage{comment}

\aclfinalcopy 

\title{Personalizing Grammatical Error Correction: \\Adaptation to Proficiency Level and L1}

\author{Maria N\u{a}dejde \\
  Grammarly \\
  \texttt{maria.nadejde@grammarly.com} 
  \And
  Joel Tetreault\thanks{This research was conducted while the author was at Grammarly.} \\
  Dataminr \\
  \texttt{jtetreault@dataminr.com}
  \\}

\date{}

\begin{document}
\maketitle
\begin{abstract}
 Grammar error correction (GEC) systems have become ubiquitous in a variety of software applications, and have started to approach human-level performance for some datasets. However, very little is known about how to efficiently personalize these systems to the user's characteristics, such as their proficiency level and first language, or to emerging domains of text. We present the first results on adapting a general purpose neural GEC system to both the proficiency level and the first language of a writer, using only a few thousand annotated sentences. Our study is the broadest of its kind, covering five proficiency levels and twelve different languages, and comparing three different adaptation scenarios: adapting to the proficiency level only, to the first language only, or to both aspects simultaneously.  We show that tailoring to both scenarios achieves the largest performance improvement (3.6 F$_{0.5}$) relative to a strong baseline.
 

\end{abstract}

\section{Introduction}
Guides for English teachers have extensively documented how grammatical errors made by learners are influenced by their native language (L1). \citet{LearnerEnglish} attribute some of the errors to ``transfer" or ``interference" between languages. For example, German native speakers are more likely to incorrectly use a definite article with general purpose nouns or omit the indefinite article when defining people's professions. Other errors are attributed to the absence of a certain linguistic feature in the native language. For example, Chinese and Russian speakers make more errors involving articles, since these languages do not have articles. 

A few grammatical error correction (GEC) systems have incorporated knowledge about L1. \citet{Rozovskaya:2011:ASM:2002472.2002589} use a different prior for each of five L1s to adapt a Naive Bayes classifier for preposition correction. \citet{rozovskaya2017adapting} expand on this work to eleven L1s and three error types.  \citet{I11-1017} showed for the first time that a statistical machine translation (SMT) system applied to GEC performs better when the training and test data have the same L1. \citet{D16-1195} extend this work by adapting a neural language model to three different L1s and use it as a feature in SMT-based GEC system. However, we are not aware of prior work addressing the impact of both proficiency level and native language on the performance of GEC systems. Furthermore, neural GEC systems, which have become state-of-the-art~\cite{gehring2017convs2s, N18-1055, N18-2046}, 
are general purpose and domain agnostic.

We believe the future of GEC lies in providing users with feedback that is personalized to their proficiency level and native language (L1). In this work, we present the first results on adapting a general purpose neural GEC system for English to both of these characteristics by using fine-tuning, a transfer learning method for neural networks, which has been extensively explored for domain adaptation of machine translation systems~\cite{Luong-Manning:iwslt15, DBLP:journals/corr/FreitagA16, P17-2061, D17-1156, thompson-EtAl:2018:WMT}. We show that a model adapted to both L1 and proficiency level outperforms models adapted to only one of these characteristics. Our contributions also include the first results on adapting GEC systems to proficiency levels and the broadest study of adapting GEC to L1 which includes twelve different languages.



\begin{figure*}[!ht]
\centering
\includegraphics[scale=0.4]{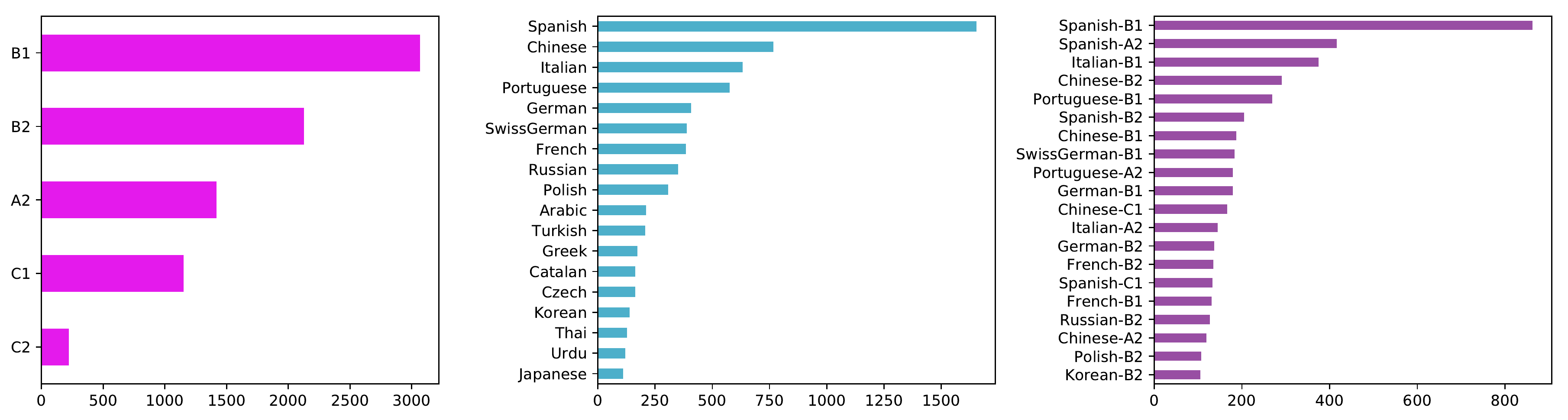} 
    \caption{Corpus Distributions for CEFR Level, L1 and L1-Level.}
    \label{fig:randbase-dist}
\end{figure*}

\section{Personalizing GEC}

\paragraph{Data}
In this work, we adapt a general purpose neural GEC system, initially trained on two million sentences written by both native and non-native speakers and covering a variety of topics and styles.  All the sentences have been corrected for grammatical errors by professional editors.\footnote{To maintain anonymity, we do not include more details.}

Adaptation of the model to proficiency level and L1 requires a corpus annotated with these features.  We use the Cambridge Learner Corpus (CLC)~\citep{nicholls2003cambridge} comprising examination essays written by English learners with six proficiency levels\footnote{The CLC uses levels defined by the Common European Framework of Reference for Languages: A1 - Beginner, A2 - Elementary, B1 - Intermediate, B2 - Upper intermediate, C1 - Advanced, C2 - Proficiency.} and more than 100 different native languages. Each essay is corrected by one annotator, who also identifies the minimal error spans and labels them using about 80 error types. From this annotated corpus we extract a parallel corpus comprising of source sentences with grammatical errors and the corresponding corrected target sentences. 

We do note the proprietary nature of the CLC which makes reproducibility difficult, though it has been used in prior research, such as \citet{rei-yannakoudakis-2016-compositional}.  It was necessary for this study as the other GEC corpora available are not annotated for both L1 and level. 
The Lang-8 Learner Corpora~\cite{I11-1017} also provides information about L1, but it has no information about proficiency levels. The FCE dataset~\citep{yannakoudakis-etal-2011-new} is a subset of the CLC, however, it only covers one proficiency level and there are not enough sentences for each L1 for our experiments. Previous work on adapting GEC classifiers to L1~\citep{rozovskaya2017adapting} used the FCE corpus, and thus did not address adaptation to different proficiency levels. One of our future goals is to create a public corpus for this type of work.  





\paragraph{Experimental Setup}

Our baseline neural GEC system is an RNN-based encoder-decoder neural network with attention and LSTM units~\citep{bahdanau2015neural}. The system takes as input an English sentence which may contain grammatical errors and decodes the corrected sentence. We train the system on the parallel corpus extracted from the CLC with the OpenNMT-py toolkit~\cite{opennmt} using the  hyper-parameters listed in the Appendix.
To increase the coverage of the neural network's vocabulary, without hurting efficiency, we break source and target words into sub-word units. The segmentation into sub-word units is learned from unlabeled data using the Byte Pair Encoding (BPE) algorithm~\cite{sennrich2015bpe}. The vocabulary, consisting of 20,000 BPE sub-units, is shared between the encoder and decoder.\footnote{Although the source and target vocabularies are the same, the embeddings are not tied.} We truncate sentences longer than 60 BPE sub-units and train the baseline system with early stopping on a development set sampled from the base dataset.\footnote{Performance did not improve after 15 epochs.}

To train and evaluate the adapted models, we extract subsets of sentences from the CLC that have been written by learners having a particular Level, L1, or L1-Level combination. We consider all subsets having at least 11,000 sentences, such that we can allocate 8,000 sentences for training, 1,000 for tuning and 2,000 for testing. We compare adapted models trained and evaluated on the same subset of the data. For example, we adapt a model using the Chinese training data and then evaluate it on the Chinese test set. 

Since our base dataset and CLC are different domains, we wanted to make sure that improvements by fine-tuning by Level or L1 were not due to simply being in-domain with the test data, which is also from the CLC.  To control for this, we construct another baseline system (``Random'') by adapting the general purpose GEC system to a random sample of learner data drawn from the CLC.
In Figure~\ref{fig:randbase-dist} we show the distribution of Level, L1 and L1-Level sentences in a random CLC sample, for the subsets having at least 100 sentences.  
B1 is the most frequent level, while A2, the lowest proficiency level included in this study, is half as frequent in the random sample. The L1 distribution is dominated by Spanish, with Chinese second with half as many sentences. Among the L1-Level subsets, Spanish-B2 is the most frequent with Spanish-A2 covering half as many sentences.

\paragraph{Fine-tuning}

We build adapted GEC models using fine-tuning, a transfer learning method for neural networks. We continue training the parameters of the general purpose model on the ``in-domain'' subset of the data covering a particular Level, L1, or L1-Level. \citet{thompson-EtAl:2018:WMT} showed that adapting only a single component of the encoder-decoder network is almost as effective as adapting the entire set of parameters. 
In this work, we fine-tune the parameters of the source embeddings and encoder, while
keeping the other parameters fixed. 

To avoid quickly over-fitting to the smaller ``in-domain'' training data, we reduce the batch size~\cite{thompson-EtAl:2018:WMT} and continue using the dropout regularization~\cite{D17-1156}. We apply dropout to all the layers and to the source words, as well as variational dropout~\citep{Gal:2016:TGA:3157096.3157211} on each step, all with probability 0.1. We also reduce the learning rate by four times and use the \texttt{start\_decay\_at} option which halves the learning rate after each epoch. Consequently, the updates become small after a few epochs. To enable the comparison between different adaptation scenarios, all fine-tuned models are trained for 10 epochs on 8,000 sentences of ``in-domain'' data.

\section{Results}

We report the results for the three adaptation scenarios: adapting to Level only, adapting to L1 only, and adapting to both L1 and Level.  
We summarize the results by showing the average M$^{2}$  F$_{0.5}$ score~\cite{dahlmeier-ng-2012-better} across all the test sets included in the respective scenario. 

We first note that the strong baseline (``Random"), which is a model adapted to a random sample of CLC , achieves improvements between 11 to 13 F$_{0.5}$ points on average on all scenarios.  While not the focus of the paper, this large improvement shows the performance gains by simply adapting to a new domain (in this case CLC data).  
Second, we note that the models adapted only by Level or by L1 are on average better than the ``Random" model by 2.1 and 2.3 F$_{0.5}$ points respectively. Finally, the models adapted to both Level and L1 outperform all others, beating the ``Random" baseline on average by 3.6 F$_{0.5}$ points. 

On all adaptation scenarios we report the performance of the single best model released by \citet{N18-1055}. Their model, which we call \textit{JD single}, was trained on English learner data of comparable size to our base dataset and optimized using the CoNLL14 training and test data.


\paragraph{Adaptation by Proficiency Level}

We adapt GEC models to five of the CEFR proficiency levels: A2, B1, B2, C1, C2. 
The results in Table~\ref{tab:adapt-Level} show that performance improves for all levels compared to the ``Random'' baseline. The largest improvement, 5.2 F$_{0.5}$ points, is achieved for A2, the lowest proficiency level. We attribute the large improvement to this level having a higher error rate, a lower lexical diversity and being less represented in the random sample on which the baseline is trained on. In contrast, for the B1 and B2 levels, the most frequent in the random sample, improvements are more modest: 0.7 and 0.2 F$_{0.5}$ points respectively. Our adapted models are better than the \textit{JD single} model on all levels, and with a large margin on the A2 and C1 levels.
\begin{table}[!h]
 \centering 
 \fontsize{10}{11}\selectfont 
 \begin{tabular}{l| r|r|r|r|r|r}
Adapt & A2 & B1 & B2 & C1 & C2 & Avg. \cr \hline \
No &  30.4 &  34.9 &  33.1 &  32.5 &  33.0 & 32.8 \cr \
Rand. &  48.4 &  47.9 &  42.5 &  41.4 &  39.2 & 43.8 \cr \
Level &   \bf{53.6} &   \bf{48.6} &   \bf{42.7} &   \bf{43.3} &   \bf{41.1} & \bf{45.9} \cr \hline
JD single & 44.1 & 47.1 & 41.7 & 37.8 & 35.0 & 44.1 \cr
    \end{tabular} 
    \caption{Adaptation to Proficiency Level in F$_{0.5}$}
    \label{tab:adapt-Level}
 \end{table}

\begin{table*}[!ht] 
 \centering 
 \fontsize{10}{12}\selectfont 
 \begin{tabular}{l| r|r|r|r|r|r|r|r|r|r|r|r|r}
Adapt & AR  & CN  & FR  & DE  & GR  & IT  & PL  & PT  & RU  & ES  & CH  & TR & Avg \cr \hline \
No &  37.5 &  36.2 &  32.7 &  31.4 &  32.7 &  29.3 &  36.0 &  31.7 &  35.8 &  32.1 &  31.1 &  35.4 & 33.5 \cr \
Random &  46.3 &  45.0 &  44.9 &  44.7 &  46.4 &  44.9 &  46.2 &  45.2 &  45.3 &  47.6 &  44.2 &  47.0 & 45.6 \cr \
L1 &  \bf{48.3} &  \bf{46.2} &  \bf{46.1} &  \bf{47.1} &  \bf{49.0} &  \bf{46.8} &  \bf{48.4} &  \bf{47.6} &  \bf{47.8} &  \bf{49.8} &  \bf{47.1} &  \bf{50.6} & \bf{47.9} \cr  \hline
JD single & 47.0  & 44.7  & 44.2  & 41.4  & 44.1  & 40.7  & 46.0  & 44.6  & 43.7  & 44.8  & 40.7  & 47.5 & 44.1 
\vspace{1em}
    \end{tabular} 

\begin{tabular}{l| r|r|r|r|r|r|r|r|r|r}
Adapt & CN-B2  & CN-C1  & FR-B1  & DE-B1  & IT-B1  & PT-B1  & ES-A2  & ES-B1  & ES-B2 & Avg. \cr \hline \
No & 36.1   & 32.5   & 31.8   & 31.2   & 28.1   & 31.4   & 28.9   & 31.9   & 33.7 & 31.8  \cr \
Random & 42.7   & 39.1   & 45.3   & 46.1   & 43.5   & 45.2   & 50.2   & 46.4   & 44.1 & 44.7 \cr \
Level & 43.4   & 41.0   & 46.5   & 46.9   & 45.3   & 46.1   & 56.6   & 47.5   & 43.7 & 46.3  \cr \
L1 & 44.1   & 40.9   & 46.5   & 48.1   & 46.5   & 46.2   & 53.8   & 47.6   & 44.4 & 46.5  \cr \
L1 \& Level &  \bf{45.5}  &  \bf{43.1}  &  \bf{48.1}  &  \bf{50.2}  &  \bf{47.3}  &  \bf{47.9}  &  \bf{58.2}  &  \bf{48.8}  &  \bf{45.6} & \bf{48.3} \cr
\hline
JD single & 43.0 & 35.8 & 46.9 & 43.8 & 41.6 & 46.7 & 43.4 & 45.0 & 41.0 & 43.0\cr
    \end{tabular} 

    \caption{Top: Adaptation to L1 Only. Bottom: Adaptation to Level and L1. Eval metric: F$_{0.5}$}
    \label{tab:adapt-L1-Level}
 \end{table*}
 
\paragraph{Adaptation by L1}

We adapt GEC models to twelve L1s: Arabic, Chinese, French, German, Greek, Italian, Polish, Portuguese, Russian, Spanish, Swiss-German  and Turkish. The results in  Table~\ref{tab:adapt-L1-Level} (top) show that all L1-adapted models are better than the baseline, with improvements ranging from 1.2 F$_{0.5}$ for Chinese and French, up to 3.6 F$_{0.5}$ for Turkish. For the languages that are less frequent in the random sample of CLC (Greek, Turkish, Arabic, Polish and Russian) we see consistent improvements of over 2 F$_{0.5}$ points.
Our adapted models are better than the \textit{JD single} model on all L1s, and with a margin larger than 5 F$_{0.5}$ points on German, Swiss-German, Italian, Greek and Spanish.

\paragraph{Adaptation by L1 and Proficiency Level}

Finally, we adapt GEC models to the following nine L1 -- Level subsets: Chinese-B2, Chinese-C1, French-B1, German-B1, Italian-B1, Portuguese-B1, Spanish-A2, Spanish-B1 and Spanish-B2. We include these subsets in our study because they meet the requirement of having at least 8,000 sentences for training.
All the models adapted to both Level and L1 outperform the models adapted to only one of these features, as shown in Table~\ref{tab:adapt-L1-Level} (bottom).
Focusing on the two levels for Chinese native speakers, we see the model adapted to C1 achieves a larger improvement over the baseline, 4.1 F$_{0.5}$ points, compared to 2.7 F$_{0.5}$ points for the B2 level. Again, this is explained by the lower frequency of the C1 level in the random sample of CLC, which is also reflected by the lowest F$_{0.5}$ score for the baseline model. Similarly, among the models adapted to different levels of Spanish native speakers, the one adapted to Spanish-A2 achieves the largest gains of 8 F$_{0.5}$ points. The Spanish-A2 testset has the highest number of errors per 100 words among all the L1-Level testsets, as shown in Table 1 in the Appendix. 
Furthermore, the A2 level is only half as frequent as the B1 level in the random sample of CLC.
Finally, our adapted models are better than the \textit{JD single} model on all L1--Level subsets, with a margin of 5 F$_{0.5}$ points on average.

\begin{table}[!h]
 \centering 
 \fontsize{10}{12}\selectfont 
 \begin{tabular}{l r|r|r}
Adapted & P & R & F0.5  \cr  \hline \
Random & 61.9 & 35.6 & 54.0 \cr \
CN-C1 & 61.1 & 37.0 & 54.1\cr \
CN-B2 & 62.4 & 37.5 & 55.1  \cr \
\ + spellcheck & 63.6 & 40.3 & 57.0  \cr 
\hline
JD single & 59.1 & 40.4 & 54.1 \cr \
JD ensemble & 63.1 & 42.6 & 57.5 \

\end{tabular}
\caption{Results on the CoNLL14 testsets for Chinese models.}
\label{tab:conll14-results}
\end{table}

\begin{table*}[!ht]
 \centering 
 \fontsize{10}{12}\selectfont 
 \begin{tabular}{l| r|r|r|r|r|r|r}
Adapt & Det & Prep & Verb & Tense & NNum & Noun & Pron \cr \hline
CN-C1 & 3.53 & 5.90 & 2.99 & 1.77 & 8.28 & 8.02 & 22.78  \cr \ 
FR-B1 & 2.34 & 1.99 & 12.54 & 5.16 & 9.16 & 3.48 & 1.13 \cr \ 
DE-B1 & 8.85 & 1.77 & 2.04 & 2.37 & 3.86 & 7.18 & 22.75  \cr \ 
IT-B1 & 2.37 & 5.32 & 12.48 & 6.74 & 4.40 & 3.29 & 8.99  \cr \ 
ES-A2 & 6.06 & 12.52 & 7.51 & 8.54 & 8.73 & 12.39 & 10.57  \cr \ 
    \end{tabular} 
    \caption{L1-Level breakdown by error type in relative improvements in $F_{0.5}$ over the ``Random'' baseline.}
    \label{tab:errtype-results}
 \end{table*}

\paragraph{CoNLL14 Evaluation}  We compare our adapted models on the CoNLL14 testset \cite{ng-EtAl:2014:W14-17} in Table~\ref{tab:conll14-results}. The model adapted to Chinese-B2 improves the most over the baseline, achieving  55.1 F$_{0.5}$. 
This result aligns with how the test set was constructed: it consists of essays written by university students, mostly Chinese native speakers. When we pre-process the evaluation set before decoding with a commercial spellchecker\footnote{Details removed for anonymity.}, our adapted model scores 57.0 which places it near other leading models, trained on a similar amount of data, such as \citet{chollampatt-ng-2018-neural} (56.52) and \citet{N18-1055}\footnote{We call their ensemble of four models with language model re-scoring \textit{JD ensemble} and their single best model without language model re-scoring \textit{JD single}} (57.53) 
even though we do not use the CoNLL14 in-domain training data. We note that the most recent state-of-the-art models~\citep{zhao-etal-naacl-2019, grundkiewicz-etal-2019-neural}, are trained on up to one hundred million additional synthetic parallel sentences, while we adapt models with only eight thousand parallel sentences.

\paragraph{Error-type Analysis}
\label{ETC}
We conclude our study by reporting improvements on the most frequent error types, excluding punctuation, spelling and orthography errors. We identify the error types in each evaluation set with Errant, a rule-based classifier~\cite{bryant-felice-briscoe:2017:Long}.
Table~\ref{tab:errtype-results} shows the results for the systems adapted to both L1 and Level that improved the most in overall $F_{0.5}$. The adapted systems consistently outperform the ``Random'' baseline on most error types. For Chinese-C1, the adapted model achieves the largest gains on pronoun (Pron) and noun number agreement errors (NNum). 
The Spanish-A2 adapted model achieves notable gains on preposition (Prep), noun and pronoun errors. Both the French-B1 and Italian-B1 adapted models gain the most on verb errors.
For German-B1, the adapted model improves the most on pronoun (Pron)  and determiner (Det) errors. The large improvement of 22.75 $F_{0.5}$ points for the pronoun category is in part an artefact of the small error counts. The adapted model corrects 35 pronouns (P=67.3) while the baseline corrects only 15 pronouns (P=46.9). We leave an in depth analysis by error type to future work. 

Below, we give an example of a confused auxiliary verb that the French-B1 adapted model corrects. The verb phrase corresponding to ``go shopping'' in French is ``faire des achats'', where the verb ``faire'' would translate to ``make/do''.

\begin{tabular}{l | p{5.5cm}}
Orig & He told me that celebrity can be bad because he can't \textit{do} shopping normally. \cr \hline 
Rand & He told me that \textit{the} celebrity can be bad because he can't \textit{do} shopping normally.  \cr \hline 
FR-B1 & He told me that celebrity can be bad because he can't \textit{go} shopping normally.  \cr  \hline  
Ref & He told me that celebrity can be bad because he can't \textit{go} shopping normally. \cr
\end{tabular}

\section{Conclusions}
We present the first results on adapting a neural GEC system to proficiency level and L1 of language learners. This is the broadest study of its kind, covering five proficiency levels and twelve different languages. While models adapted to either proficiency level or L1 are on average better than the baseline by over 2 F$_{0.5}$ points and the largest improvement (3.6 F$_{0.5}$) is achieved when adapting to both characteristics simultaneously.

We envision building a single model that combines knowledge across L1s and proficiency levels using a  mixture-of-experts approach.  Adapted models could also be improved by using the \textit{mixed fine tuning} approach which uses a mix of in-domain and out-of-domain data~\cite{P17-2061}.  

\section*{Acknowledgements}
The authors would like to thank the anonymous reviewers for their feedback. We are also grateful to our colleagues for their assistance and insights: Dimitrios Alikaniotis, Claudia Leacock, Dmitry Lider, Courtney Napoles, Jimmy Nguyen, Vipul Raheja and Igor Tytyk.

\bibliography{emnlp-ijcnlp-2019}
\bibliographystyle{acl_natbib}

\end{document}



\appendix
\section*{Supplemental Material}
In this appendix we cover: a) hyper-parameters, b) error frequencies c) precision and recall results for the three main adaptation experiments.


\section{Hyper-parameters}
\label{sec:appendix-hyperparams}
For pre-training the general purpose model:
\begin{verbatim}
        -epochs         15
        -word_vec_size  500
        -encoder_type   brnn
        -decoder_type   rnn
        -enc_layers     3
        -dec_layers     3
        -batch_size     296
        -rnn_size       500
        -dropout        0.1
        -learning_rate  0.001
        -optim          adam
        -start_decay_at 6
        -report_every   200
        -max_grad_norm  1.0
\end{verbatim}

For fine-tuning the models on ``in-domain'' data:
\begin{verbatim}
        -epochs         25
        -word_vec_size  500
        -encoder_type   brnn
        -decoder_type   rnn
        -feat_merge     concat
        -enc_layers     3
        -dec_layers     3
        -batch_size     128 
        -rnn_size       500
        -dropout 0.1
        -learning_rate  0.00025
        -optim          adam
        -start_decay_at 16
        -report_every   50
        -max_grad_norm  1.0
\end{verbatim}

\newpage
\section{Error Frequencies}

\begin{table}[h]
 \centering 
 \fontsize{10}{12}\selectfont 
 \begin{tabular}{l |c}
 L1-Level & \# errors \cr \hline\
CN-C1 &  12.13 \cr \ 
FR-B1 & 13.17 \cr \ 
DE-B1 &  12.48 \cr \ 
IT-B1 & 12.13 \cr \ 
PT-B1 & 13.14 \cr \ 
ES-A2 & 17.33 \cr \ 
ES-B1 & 13.28 \cr \ 
ES-B2 & 12.53 \ 
    \end{tabular} 
    \caption{The numbers of errors per 100 words for the L1-Level testsets.}
    \label{tab:errfreq}
 \end{table}

\section{Precision and Recall Results}

\begin{table*}[htbp] 
 \centering 
 \fontsize{10}{12}\selectfont 
 \begin{tabular}{l r|r|r||r|r|r||r|r|r}
 & \multicolumn{3}{c||}{A2}  & \multicolumn{3}{c||}{B1}  & \multicolumn{3}{c}{B2} \cr \
 & P & R & F0.5 & P & R & F0.5 & P & R & F0.5 \cr \hline \
None &  34.9 & 20.1 & 30.4 &  39.0 & 24.5 & 34.9 &  38.1 & 21.6 & 33.1 \cr \
Random &  57.3 & 29.9 & 48.4 &  57.5 & 28.8 & 47.9 &  51.4 & 25.1 & 42.5 \cr \
Adapted &  \bf{60.4} & \bf{37.0} & \bf{53.6} &  \bf{58.1} & \bf{29.4} & \bf{48.6} &  \bf{51.9} & \bf{24.9} & \bf{42.7} \cr \
\\
\end{tabular}

\begin{tabular}{l r|r|r||r|r|r}
& \multicolumn{3}{c}{C1}  & \multicolumn{3}{c}{C2}  \cr \
 & P & R & F0.5 & P & R & F0.5 \cr  \hline \
None &  37.8 & 20.9 & 32.5 &  39.0 & 20.5 & 33.0   \cr  \
Random &  51.7 & 23.0 & 41.4 &  50.8 & 20.5 & 39.2  \cr  \
Adapted &  \bf{54.3} & \bf{23.9} & \bf{43.3} &  \bf{54.5} & \bf{20.8} & \bf{41.1}  \cr  \
   \end{tabular} 
       \caption{Adaptation by proficiency level.}
 \end{table*}

\begin{table*}[!ht]
\centering
\begin{tabular}{l r|r|r||r|r|r||r|r|r||r|r|r}
  & \multicolumn{3}{c||}{Chinese}  & \multicolumn{3}{c||}{Italian}  & \multicolumn{3}{c||}{Portuguese}  & \multicolumn{3}{c}{Spanish}  \cr \
 & P & R & F0.5 & P & R & F0.5 & P & R & F0.5 & P & R & F0.5 \cr  \hline \
None &  42.6 & 22.7 & 36.2 &  33.4 & 19.6 & 29.3 &  36.9 & 20.2 & 31.7 &  36.5 & 21.6 & 32.1 \cr \
Random &  54.3 & 26.8 & 45.0 &  55.2 & 25.7 & 44.9 &  55.3 & 26.2 & 45.2 &  57.5 & 28.2 & 47.6 \cr \
Adapted &  \bf{54.4} & \bf{28.9} & \bf{46.2} &  \bf{57.5} & \bf{26.8} & \bf{46.8} &  \bf{56.9} & \bf{28.8} & \bf{47.6} &  \bf{59.5} & \bf{30.1} & \bf{49.8} \cr \
\\

& \multicolumn{3}{c||}{French}  & \multicolumn{3}{c||}{German}  & \multicolumn{3}{c||}{Russian}  & \multicolumn{3}{c}{SwissGerman}  \cr \
 & P & R & F0.5 & P & R & F0.5 & P & R & F0.5 & P & R & F0.5 \cr \hline
 \
None &  37.8 & 21.2 & 32.7 &  35.1 & 22.2 & 31.4 &  41.1 & 23.8 & 35.8 &  35.4 & 20.9 & 31.1 \cr \
Random &  54.9 & 25.9 & 44.9 &  55.0 & 25.5 & 44.7 &  54.0 & 27.6 & 45.3 &  54.4 & 25.2 & 44.2 \cr \
Adapted &  \bf{56.8} & \bf{26.3} & \bf{46.1} &  \bf{58.3} & \bf{26.7} & \bf{47.1} &  \bf{56.7} & \bf{29.5} & \bf{47.8} &  \bf{57.8} & \bf{27.2} & \bf{47.1} \cr \

\\

& \multicolumn{3}{c||}{Arabic}  & \multicolumn{3}{c||}{Greek}  & \multicolumn{3}{c||}{Polish}  & \multicolumn{3}{c}{Turkish}  \cr \
 & P & R & F0.5 & P & R & F0.5 & P & R & F0.5 & P & R & F0.5 \cr  \hline \
None &  43.0 & 24.9 & 37.5 &  36.4 & 23.1 & 32.7 &  40.3 & 25.2 & 36.0 &  40.0 & 24.4 & 35.4 \cr \
Random &  54.9 & 28.5 & 46.3 &  56.3 & 27.2 & 46.4 &  54.5 & 28.7 & 46.2 &  55.7 & 29.0 & 47.0 \cr \
Adapted &  \bf{55.8} & \bf{31.3} & \bf{48.3} &  \bf{58.4} & \bf{29.8} & \bf{49.0} &  \bf{57.0} & \bf{30.2} & \bf{48.4} &  \bf{58.7} & \bf{32.7} & \bf{50.6} \cr \

    \end{tabular} 
    \caption{Adaptation by L1}
 \end{table*}

\begin{table*}[!ht] 
 \centering 
 \fontsize{10}{12}\selectfont 
 \begin{tabular}{l r|r|r||r|r|r||r|r|r}
 & \multicolumn{3}{c||}{Chinese-B2} & \multicolumn{3}{c||}{Chinese-C1} & \multicolumn{3}{c}{French-B1} \cr \
 & P & R & F0.5 & P & R & F0.5 & P & R & F0.5 \cr \hline \
None   & 41.4 & 23.9 & 36.1   & 39.9 & 18.6 & 32.5   & 36.4 & 21.0 & 31.8  \cr \
Random   & 51.2 & 25.6 & 42.7   & 49.9 & 20.9 & 39.1   & 54.8 & 26.7 & 45.3  \cr  \
Adapted Level  & 51.9 & 26.1 & 43.4  & 52.2 & 22.0 & 41.0  & 55.7 & 27.9 & 46.5 \cr \
Adapted L1  & 52.1 & 27.4 & 44.1  & 51.3 & 22.6 & 40.9  & 56.4 & 27.2 & 46.5 \cr \
Adapted L1 \& Level  & \bf{53.5} & \bf{28.4} & \bf{45.5}  & \bf{52.9} & \bf{24.8} & \bf{43.1}  & \bf{57.6} & \bf{29.0} & \bf{48.1} \cr \
    \end{tabular} 
    
    \\
    
    \begin{tabular}{l r|r|r||r|r|r||r|r|r}
 & \multicolumn{3}{c||}{German-B1} & \multicolumn{3}{c||}{Italian-B1} & \multicolumn{3}{c}{Portuguese-B1} \cr \
 & P & R & F0.5 & P & R & F0.5 & P & R & F0.5 \cr \hline \
None   & 35.3 & 21.2 & 31.2   & 32.1 & 18.8 & 28.1   & 36.2 & 20.6 & 31.4  \cr \
Random   & 56.5 & 26.5 & 46.1   & 54.7 & 24.0 & 43.5   & 55.1 & 26.2 & 45.2  \cr \
Adapted Level  & 57.0 & 27.4 & 46.9  & 56.4 & 25.3 & 45.3  & 56.0 & 27.0 & 46.1 \cr \
Adapted L1  & 59.2 & 27.5 & 48.1  & 58.6 & 25.5 & 46.5  & 55.2 & 28.0 & 46.2 \cr \
Adapted L1 \& Level  & \bf{60.9} & \bf{29.5} & \bf{50.2}  & \bf{58.6} & \bf{26.6} & \bf{47.3}  & \bf{57.5} & \bf{28.7} & \bf{47.9} \cr \
    \end{tabular} 
    
    \\
    
    \begin{tabular}{l r|r|r||r|r|r||r|r|r}
 & \multicolumn{3}{c||}{Spanish-A2} & \multicolumn{3}{c||}{Spanish-B1} & \multicolumn{3}{c}{Spanish-B2} \cr \
 & P & R & F0.5 & P & R & F0.5 & P & R & F0.5 \cr \hline \
None   & 32.8 & 19.7 & 28.9   & 35.8 & 22.1 & 31.9   & 38.9 & 22.1 & 33.7  \cr \
Random   & 58.7 & 31.8 & 50.2   & 55.6 & 27.9 & 46.4   & 54.4 & 25.1 & 44.1  \cr \
Adapted Level  & 62.7 & 40.8 & 56.6  & 56.8 & 28.8 & 47.5  & 54.0 & 24.8 & 43.7 \cr \
Adapted L1  & 61.3 & 36.1 & 53.8  & 56.4 & 29.2 & 47.6  & 54.4 & 25.6 & 44.4 \cr \
Adapted L1 \& level  & \bf{63.7} & \bf{43.2} & \bf{58.2}  & \bf{57.5} & \bf{30.3} & \bf{48.8}  & \bf{56.0} & \bf{26.1} & \bf{45.6} \cr \
    \end{tabular} 
        \caption{Adaptation by L1 and proficiency level. Comparison with adaptation by Level only or by L1 only.}
 \end{table*}